%% file: main.tex
\documentclass[10pt,twocolumn,letterpaper]{article}

\usepackage{iccv}      % To produce the REVIEW version
\iccvfinalcopy

\usepackage{graphicx}
\usepackage{amsmath}
\usepackage{amssymb}
\usepackage{booktabs}
\usepackage{units} 
\usepackage{times}
\usepackage{tabularx}
\usepackage{nicefrac}
\usepackage{enumitem}

\usepackage{caption}
\usepackage{xcolor}
\usepackage{mathtools}
\usepackage{xfrac}
\usepackage{multirow}
\usepackage{wasysym}
\usepackage{algorithm}% http://ctan.org/pkg/algorithm
\usepackage{algpseudocode}% http://ctan.org/pkg/algorithmicx

\makeatletter

\@namedef{ver@everyshi.sty}{}
\newcommand\notsotiny{\@setfontsize\notsotiny{6.31415}{7.1828}} % https://tex.stackexchange.com/questions/432675/textsize-between-scriptsize-and-tiny

\makeatother

\usepackage{pgfplots}
\usepackage{tikz} 
\usetikzlibrary{patterns}
\pgfplotsset{compat=newest}
\usepackage{numprint}
\usepackage[normalem]{ulem}

\usepackage[pagebackref,breaklinks,colorlinks]{hyperref}

\usepackage[capitalize]{cleveref}
\crefname{section}{Sec.}{Secs.}
\Crefname{section}{Section}{Sections}
\Crefname{table}{Table}{Tables}
\crefname{table}{Tab.}{Tabs.}

\newcommand{\probP}{\kern0.15em \text{I\kern-0.15em P}}

\newcommand{\wet}{\textvisiblespace \kern0.25em (ours)}

\newcommand{\both}{\textvisiblespace \kern-0.45em * (ours)}

\DeclareMathOperator{\EX}{\mathbb{E}}% expected value

\definecolor{water_color}{RGB}{69,156,238}
\definecolor{eth_orange}{RGB}{255,126,40}
\definecolor{eth_green}{RGB}{153,204,51}
\definecolor{eth_blue}{RGB}{169,204,242}
\definecolor{eth_red}{RGB}{230,140,132}
\definecolor{eth_gray}{RGB}{126,126,126}

\newcommand{\PAR}[1]{\vspace{-0.2eM}\vskip4pt \noindent{\bf #1}}

\def\iccvPaperID{6275} % *** Enter the ICCV Paper ID here

\begin{document}

\title{\vspace{-10pt}ScatterNeRF: \\Seeing Through Fog with Physically-Based Inverse Neural Rendering}

\author{\centerline{\hspace{-0.5eM}Andrea Ramazzina$^1$ Mario Bijelic$^2$ Stefanie Walz$^1$ Alessandro Sanvito$^1$ Dominik Scheuble$^1$ Felix Heide$^{2,3}$}
	\and
	\centerline{$^1$Mercedes-Benz \quad $^2$Princeton University \quad $^3$Algolux}}
\maketitle

%%%%%%%%% ABSTRACT
\input{sections/abstract}

%%%%%%%%% BODY TEXT
\input{sections/introduction}
%------------------------------------------------------------------------
\input{sections/related_work}
%------------------------------------------------------------------------
\input{sections/method}
%------------------------------------------------------------------------
\input{sections/results}
%------------------------------------------------------------------------
\input{sections/conclusion}

%%%%%%%%% REFERENCES
{\small
\bibliographystyle{ieee_fullname}
\interlinepenalty=10000
\bibliography{refs}
}

\end{document}

%% file: sections/abstract.tex
\begin{abstract}
\vspace{-0.7eM}
Vision in adverse weather conditions, whether it be snow, rain, or fog is challenging. In these scenarios, scattering and attenuation severly degrades image quality. Handling such inclement weather conditions, however, is essential to operate autonomous vehicles, drones and robotic applications where human performance is impeded the most. 
A large body of work explores removing weather-induced image degradations with dehazing methods. Most methods rely on single images as input and struggle to generalize from synthetic fully-supervised training approaches or to generate high fidelity results from unpaired real-world datasets. 
With data as bottleneck and most of today's training data relying on good weather conditions with inclement weather as outlier, we rely on an inverse rendering approach to reconstruct the scene content.
We introduce ScatterNeRF, a neural rendering method which adequately renders foggy scenes and decomposes the fog-free background from the participating media -- exploiting the multiple views from a short automotive sequence without the need for a large training data corpus. Instead, the rendering approach is optimized on the multi-view scene itself, which can be typically captured by an autonomous vehicle, robot or drone during operation. Specifically, we propose a disentangled representation for the scattering volume and the scene objects, and learn the scene reconstruction with physics-inspired losses. We validate our method by capturing multi-view In-the-Wild data and controlled captures in a large-scale fog chamber. %Both data and code will be published. 
\end{abstract}

%% file: sections/introduction.tex
\section{Introduction}
Imaging and scene understanding in the presence of scattering media, such as fog, smog, light rain and snow, is an open challenge for computer vision and photography. As rare out-of-distribution events that occur based on geography and region~\cite{STF}, these weather phenomena can drastically reduce the image quality of the captured intensity images, reducing local contrast, color reproduction, and image resolution~\cite{STF}. A large body of existing work has investigated methods for dehazing~\cite{shi2021zeroscatter, PFF, EPDN, zerorestore, d4Unpaired, CLAHE} with the most successful methods employing learned feed-forward models~\cite{shi2021zeroscatter, PFF, EPDN, zerorestore, d4Unpaired}. Some methods \cite{EPDN,PFF, li2017aod} use synthetic data and full supervision, but struggle to overcome the domain gap between simulation and real world. Acquiring paired data in real world conditions is challenging 
and existing methods either learn natural image priors from large unpaired datasets~\cite{Yang_2022_CVPR, d4Unpaired}, or they rely on cross-modal semi-supervision to learn to separate atmospheric effects from clear RGB intensity~\cite{shi2021zeroscatter}. Unfortunately, as the semi-supervised training cues are weak compared to paired supervised data, these methods often fail to completely separate atmospheric scatter from clear image content, especially at long distances. The problem of predicting clear images in the presence of haze is an open challenge, and notably harsh weather also results in severely impaired human vision -- a major driver behind fatal automotive accidents~\cite{DrivingBlindWeatherRelatedVisionHazardsandFatalMotorVehicleCrashes}.

\begin{figure}[t!]
	\centering
	\includegraphics[width=0.48\textwidth]{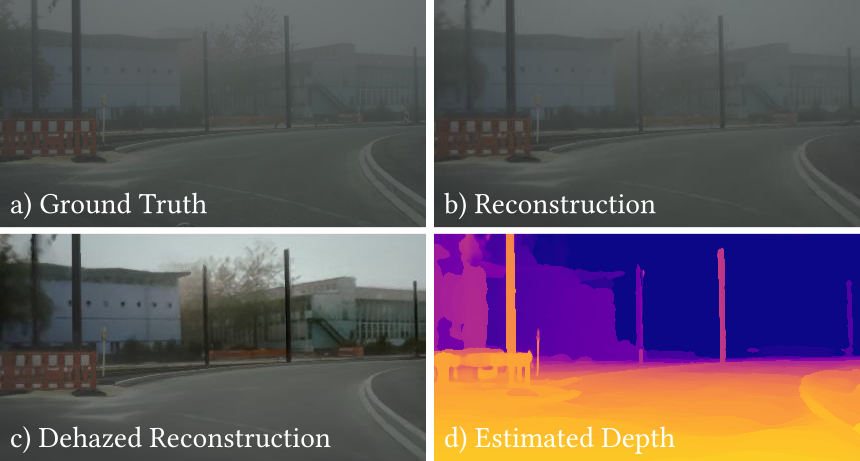}   

\vspace*{-1mm}
	\caption{ScatterNeRF produces accurate renderings for scenes with volumetric scattering (b). By learning a disentangled representation of participating media and clear scene, the proposed method is able to recover dehazed scene content (c) with accurate depth (d).}
	\label{fig:teaser}
	\vspace*{-3mm}
\end{figure}
As the distribution of natural scenes with participating media is long-tailed in typical training datasets~\cite{Ettinger_2021_ICCV, Sun_2020_CVPR, Geiger2012CVPR, Geiger2013IJRR,STF}, this also makes training and evaluation of computer vision tasks that operate on RGB streams in bad weather challenging. For supervised approaches to scene understanding tasks, these ''edge'' scenarios often directly result in failure cases, including detection~\cite{STF}, depth estimation~\cite{depthbenchmark2019}, and segmentation~\cite{SDV18}. To tackle weather-based dataset bias, existing methods have proposed augmentation approaches that either simulate atmospheric weather effects on clear images \cite{SDV18, Tremblay} or they employ fully synthetic simulation to generate physically-based adverse weather scenarios~\cite{Dosovitskiy17, halder2019physics, SDV18}. Unfortunately, both directions cannot compete with supervised training data, either due to the domain gap between real and synthetic data, or, as a result of an approximate physical forward model~\cite{Tremblay}. \newline
As such, the capability of both physically accurate modeling and separating scattering in participating media is essential for imaging and scene understanding tasks. 

In this work, we depart from both feed-forward dehazing methods and fully synthetic training data, and we address this challenge as an inverse rendering problem. Instead of predicting clean images from RGB frames, we propose to learn a neural scene representation that explains foggy images with a physically accurate forward rendering process. Once this representation is fitted to a scene, this allows us to render \emph{novel views with real-world physics-based scattering}, \emph{disentangle appearance and geometry without scattering} (i.e., reconstruct the dehazed scene), and \emph{estimate physical scattering parameters} accurately. To be able to optimize a scene representation efficiently, we build on the large body of neural radiance field methods~\cite{mildenhall2021nerf,yuan2022nerfEditing, wang2022clipNeRF, wu2022object, xu2022deforming, ost2021NSG, kundu2022panoptic,boss2021nerd, rudnev2022nerfosr, srinivasan2021nerv, boss2022samurai,zhang2020nerf++, barron2022mip, turki2022mega}. While existing NeRF methods assume peaky ray termination distributions and free-space propagation, we propose a forward model that can accurately model participating media. As an inductive bias, the scene representation, by design, separates learning the clear 3D scene and the participating media. We validate that the proposed method accurately models foggy scenes in real-world and controlled scenes, and we demonstrate that the disentangled scene intensity and depth outperform existing dehazing and depth estimation methods in diverse driving scenes.

Specifically, we make the following contributions:
\begin{itemize}
	\itemsep-0.3em

  \item We propose a method to learn a disentangled representation of the participating media by introducing the Koschmieder scattering model into the volume rendering process.
\item Our approach adds a single MLP used to model the scattering media proprieties and does not require any additional sampling or other procedures, making it a lightweight framework in terms of both computation and memory consumption.
  \item We validate that our method learns a physics-based representation of the scene and enables control over its hazed appearance. We confirm this using real data captured in both controlled and in-the-wild settings.
\end{itemize}

%% file: sections/related_work.tex
\section{Related Work}
\PAR{Imaging and Vision in Participating Media.}
%Imaging and vision in participating media 
As real world data in participating media is challenging to capture \cite{STF,li2019benchmarking,ancuti2018haze,Ancuti2018a}, a large body of work introduces simulation techniques for snowfall \cite{liu2018desnownet}, rainfall \cite{hasirlioglu2019rainsimulation,halder2019physics}, raindrops on the windshield \cite{Bernuth2018}, blur \cite{kupyn2018deblurgan}, night \cite{Sakaridis2019} and fog \cite{shi2021zeroscatter, SDV18, Li2017a, Galdran2018}. Using this data, existing methods investigate pre-processing approaches using dehazing \cite{shi2021zeroscatter, PFF, EPDN, zerorestore, d4Unpaired, CLAHE, he2010darkchannelprior,li2017aod}, deraining \cite{Hu_2019_CVPR,Clearing_the_skies} and desnowing \cite{liu2018desnownet}.
Early works on image dehazing as \cite{he2010darkchannelprior, tang2014investigating} explore image statistics and physical models to estimate the airlight introduced by the fog scattering volume and invert the Koschmieder fog model. Later, CNN approaches \cite{li2017aod} and \cite{zerorestore} learn to predict the airlight and transmission, with the same goal of inverting the Koschmieder model. However, this disjoint optimization can lead to error accumulation. Hence, \cite{EPDN, PFF, shi2021zeroscatter, d4Unpaired,Guo_2022_CVPR} model the fog removal through a neural network learned end-to-end. As such, existing methods differ substantially in network structure and learning methodology.
For example, some methods rely on GAN architectures \cite{EPDN,shi2021zeroscatter}, transformer-based backbones \cite{Guo_2022_CVPR} or encoder-decoder structures \cite{PFF}. 
For model-learning, the approaches apply semi-/self-supervised \cite{d4Unpaired,shi2021zeroscatter}, fully supervised \cite{EPDN,PFF} and test time optimization techniques \cite{MultiDomainTestTimeDehazing,zerorestore}. 
All of these methods have in common that they do not explore multiple views to reconstruct a clear image. To overcome this methodological limitation we introduce a novel multi-view dataset in hazy conditions in Sec.~\ref{sec:dataset} and explore reconstruction from multiple views for optimal image reconstruction through neural rendering approaches. Most similar to our method is the approach from Sethuraman et al.~\cite{sethuraman2022waternerf} that reconstructs scenes in underwater conditions which requires tackling strong color aberrations and specular reflections specific to the underwater domain. 

\begin{figure*}[!t]
\vspace{-6mm}
    \centering
    \includegraphics[width=1\textwidth]{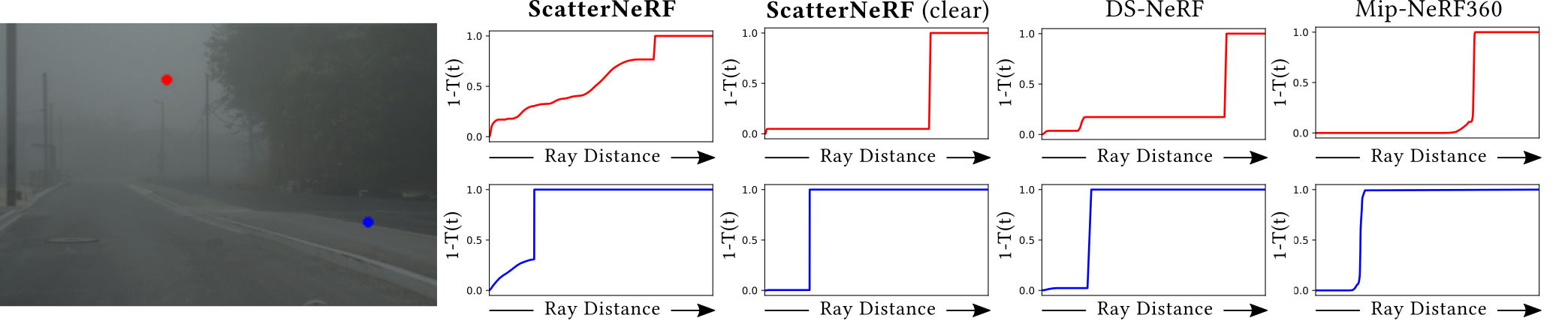}
    \vspace{-2eM}
    \caption{
    We show the ray termination distribution along two cast rays for our approach and two references NeRF models. The regularization methods proposed in DS-NeRF \cite{deng2022depth} and mip-NeRF-360 \cite{barron2022mip} represent the accumulated transmittance $T$ as a step function, whereas ScatterNeRF models the scattering process.}
    
    \label{fig:arch}
    \vspace{-3mm}
\end{figure*}

\PAR{Neural Rendering Methods for Large Scale Scenes}
%Neural fields 
A rapidly growing body of work is capable of representing unbounded outdoor scenes, such as the ones tackled in this work, with rich detail both in close and far range. NeRF++ \cite{zhang2020nerf++} achieves this by using two NeRFs, one to model the foreground and one for the background scene. DONeRF \cite{neff2021donerf} warps the space by radial distortion to bring the background closer. Recently, \cite{barron2022mip} has tackled this problem by proposing a non-linear parametrization technique suitable for the mip-NeRF algorithm \cite{barron2021mip} .
For very large-scale scenes, too big to be fitted by a single MLP, several works \cite{tancik2022block} \cite{turki2022mega} have explored the idea of learning multiple NeRFs for subparts of the scene.

\PAR{Clear Scene Priors}
Given the under-constrained nature of scene reconstruction from a sparse set of views, novel views rendered by NeRFs are frequently afflicted by floating artifacts \cite{barron2022mip, deng2022depth} or not able to properly generalize to novel views and hence fail to render images from unseen poses correctly \cite{kim2022infonerf, jain2021nerfdiet}. 
To tackle this issue, several works have recently proposed the introduction of different regularization techniques.
In \cite{kim2022infonerf}, an entropy-based loss is introduced in order to enforce a sparsity constraint over the scene. Analogously, recent models exploiting an estimated depth as training cue \cite{deng2022depth, roessle2022dense} implicitly enforce a sparsity constrain of the ray termination distributions by penalizing non-zero probabilities lying far from the prior estimated depth. Mip-NeRF360 \cite{barron2022mip} relies on a regularization technique aimed at encouraging a unimodal peaky distribution for the termination probability of a ray. \newline
While such methods work well for clear scenes, they are based on the assumption that most of the scene density is null, except for where there are solid objects. As such prior is not applicable in the presence of a participating media in the air, such approaches are not suitable for scenes with scattering media.

We propose to learn a separate representation of the clear scene and the participating media. This allows us to potentially use any of the above-mentioned regularization technique on the clear scene model without compromising on the hazed scene reconstruction.

%% file: sections/method.tex
\section{Disentangled Scattering Neural Radiance Fields}
ScatterNeRF has five integral parts, namely the underlying physical model in Sec.~\ref{sec:Kohnschmieder}, the formulation of the neural radiance field in Sec.~\ref{sec:NeuralRadienceField}, the formulation of the loss functions in Sec.~\ref{sec:TraingSupervision}, the details on ray sampling in Sec.~\ref{sec:Sampling} and implementation in Sec.~\ref{sec:Implementation}. We describe these components in the following.

\subsection{Physical Scattering Model}\label{sec:Kohnschmieder}
Large scattering volumes can be approximated by the Koschmieder model \cite{koschmieder1924theorie}. For each pixel, we model the attenuation representing the lost intensity due to scattered direct light of an object and the contribution of the airlight $\mathbf{c}_p$ caused by ambient light scattered towards the observer,
\begin{equation}
\label{eqn:koschmiederClassic}
\mathbf{C}_F = l\mathbf{C}_c + (1-l)\mathbf{C}_p,
\end{equation}
with $l$ being the transmission, $\mathbf{C}_F$ corresponding to the observed pixel value and $\mathbf{C}_c$ the clear scene. 

The transmission $l$ can be computed from the attenuation coefficient $\sigma_p$ and the depth $D$,
\begin{equation}
l = \exp(-\sigma_p D).
\end{equation}
In existing methods \cite{SDV18} both parameters $\sigma_p$ and airlight $\mathbf{c}_p$ are globally constant.
Koschmieder's model is equivalent to a volume rendering NeRF~\cite{mildenhall2021nerf}, Eq. \eqref{eq:NerfBase}, with scattering density $\sigma_p$ and airlight $\mathbf{C}_p$ set constant, and ray integration from $t_n$ to a maximum distance $t_f$. For simplicity of notation, we omit the viewing direction \(\textbf{d}\). Starting with the forward model 
\begin{equation}\label{eq:NerfBase}
\mathbf{C}_F(\mathbf{r}) = \int_{t_n}^{t_f} T_F(t) \sigma_F(\mathbf{r}(t)) \mathbf{c}_F(\mathbf{r}(t)) dt, \ 
\end{equation}
and assuming two disjoint additive volume densities  $\sigma_c$ for the scene and $\sigma_p$ for the scattering media ($\sigma_F=\sigma_p+\sigma_c$), the volume rendering equation can be formulated as
\begin{equation}
\label{eqn:koschmiederNeRF}
\mathbf{C}_{F}(\mathbf{r}) = l(\mathbf{r}) \mathbf{C}_{c}(\mathbf{r}) +
\left(1-  l(\mathbf{r})   \right) \mathbf{C}_{p},
\end{equation}
 where the integrated object color $\mathbf{C}_c$ is defined with emitted color $\mathbf{c}_c(\mathbf{r})$ at position $\mathbf{r}$,
\begin{equation}
\mathbf{C}_{c}(\mathbf{r}) = \int_{D}^{t_f} T_c(t) \sigma_c(\mathbf{r}(t)) c_c(\mathbf{r}(t)) dt.
 \end{equation}
The transmission then is dependent on $\mathbf{r}$ leading to $\exp\left( -\sigma_{p} D \right)$.
Here, the subscript denotes if the clear scene object $c$ or participating media $p$ is modeled.
Finally, the value of the foggy pixel $\mathbf{C}_F(\mathbf{r})$ can be estimated by integrating $\mathbf{c}_F(\mathbf{r}(t))$ along one camera ray vector $\mathbf{r}$. We denote $\mathbf{r}_i = [\mathbf{x}_i;\mathbf{d}_i] \in \mathbb{R}^5 $ consisting of the 3D position $\mathbf{x}$ and direction $\mathbf{d}$. \newline

Next, we can relax the constraints on $\sigma_p$ and $\mathbf{c}_p$, and allow them to approximate arbitrary values. This results in,
\begin{equation}
\label{eqn:fog_volumetric_rendering}
\begin{split}
\mathbf{C}_F(\mathbf{r}) = \int_{t_n}^{t_f} T_F(t) & ( \sigma_p(\mathbf{r}(t)) \mathbf{c}_p(\mathbf{r}(t)) \\
&+ \sigma_c(\mathbf{r}(t)) \mathbf{c}_c(\mathbf{r}(t)) )  dt,
\end{split}
\end{equation}
with,
\begin{align}
\small
T_F(t) =\ & \exp \left( -\int_{t_n}^{t} (\sigma_{c}(\mathbf{r}(s))+\sigma_{p}(\mathbf{r}(s))) ds \right),\\
%T_F(t) =\ & \exp \left( -\int_{t_n}^{t} (\sigma_{c}(\mathbf{r}(s)) ds\right)\exp\left(-\int_{t_n}^{t}\sigma_{p}(\mathbf{r}(s))) ds \right), \\
T_F(t) =\ & T_p(t)T_c(t),
\end{align}
which can be expressed as
\begin{align}
%\begin{split}
\mathbf{C}_F(\mathbf{r}) =\ & \int_{t_n}^{t_f} T_c(t) \underbrace{\left(T_p(t) \sigma_p(\mathbf{r}(t)) \mathbf{c}_p(\mathbf{r}(t)) \right)}_{\text{Fog Contribution}} \nonumber
\\
& \quad + T_p(t)\underbrace{\left(T_c(t)\sigma_c(\mathbf{r}(t)) \mathbf{c}_c(\mathbf{r}(t)) \right) }_{\text{Clear scene contribution}}dt. \label{eq:FinalFogNerfBase}
\end{align}
\subsection{Neural Radiance Model}\label{sec:NeuralRadienceField}
Further simplifications of Eq.~\eqref{eq:NerfBase} can be found by solving the integral through numerical quadrature needed for the discrete neural forward network.
The numerical quadrature leads to,
\begin{equation}
    \mathbf{C}_F(\mathbf{r}) = \sum_{i}^{N} w_{F}(\mathbf{r}(t_i)) \mathbf{c}_{F}(\mathbf{r}(t_i)). \label{eq:QuadratureBase}
\end{equation}
\begin{align}
        w_{F}(\mathbf{r}_i) =\ & T_{F}(\mathbf{r}_i)  (1-\exp\left(\left(\sigma_{p}(\mathbf{r}_i) +\sigma_{c}(\mathbf{r}_i)\right)\delta_j \right), \\
        T_{F}(\mathbf{r}_i) =\ & \exp \left( -\sum_{j=1}^{i-1} \left(\sigma_{p}(\mathbf{r}_i) +\sigma_{c}(\mathbf{r}_i)\right) \delta_i \right),\\
        \delta_i =\ & t_{t+1} - t_i,\ \text{and} \\
        \mathbf{c}_{F}(\mathbf{r}_i, \mathbf{d}) =\ & \frac{\sigma_{c}(\mathbf{r}_i) \mathbf{c}_{c}(\mathbf{r}_i) + \sigma_{p}(\mathbf{r}_i) \mathbf{c}_{p}(\mathbf{r}_i) }{\sigma_{p}(\mathbf{r}_i) +\sigma_{c}(\mathbf{r}_i)}.
\end{align}
To facilitate the learning of two independent volume representations, we model each part independently by one NeRF. The parameters $\sigma_i, \mathbf{c}_i$ for $i\ \in\ \lbrace c,p \rbrace$ are predicted by a multi-layer perceptron (MLP) as,
\begin{equation}
    {\mathbf{c}_i,\sigma_i} = f_i \left( \gamma(\mathbf{x})\right).
\end{equation}\label{eqn:volumerendering}
For the clear scene NeRF we adopt a similar strategy as in \cite{mildenhall2021nerf} and optimize simultaneously two MLPs, $f_{c_{coarse}}$ and $f_{c_{fine}}$ using the same loss formulation but different sampling procedure, as detailed in Sec.~\ref{sec:TraingSupervision} and Sec.~\ref{sec:Sampling}.
The coordinates are encoded by the function $\gamma$ following \cite{mildenhall2021nerf, tancik2020fourier}. 
This learning disentanglement of the scene and scatter representation allows us to render scenes with different fog densities by scaling $\sigma_p$ or even dehaze the image entirely by setting $\sigma_p=0$. \newline
For the dehazing task the image can be rerendered with $\sigma_p=0$ and the formulation reduces to the scene model only leading to, 
\begin{align}
    \mathbf{C}_F (\mathbf{r}) = \mathbf{C}_c (\mathbf{r}) =\ & \sum_i^N w_c(\mathbf{r}(t_i))\mathbf{c}_{c}(\mathbf{r}(t_i)), \\
    w_{c}(\mathbf{r}(t_i)) =\ & T_{c}(t_i)  (1-\exp\left(\sigma_{c}(\mathbf{r}(t_i))\delta_i \right), \\
    T_{c}(t_i) =\ & \exp \left( -\sum_{j=1}^{i-1}(\sigma_{c}(\mathbf{r}(t_j)) \delta_j \right).
\end{align}

\subsection{Training Supervision}\label{sec:TraingSupervision}
To learn the neural forward model we supervise the reconstructed images with a pixel loss between predicted and training frames $\mathcal{L}_{rgb}$, align the observable airlight $\mathcal{L}_A$, supervise the scene depth with $\mathcal{L}_D$, enforce discrete clear scene volumetric density $\mathcal{L}_{ec}$ and enforce the scattering density to be disjoint from the scene objects in $\mathcal{L}_{ep}$. In the following, network predictions are marked with a hat and ground-truth values are marked with a bar.

\PAR{Color Supervision} 
The image loss between ground-truth $\mathbf{\bar{C}}_F$ and reconstructed haze images $\mathbf{\hat{C}}_F$ is used for direct supervision as
\begin{equation}
L_{rgbF} =\EX_\mathbf{r} \left[ || \mathbf{\hat{C}}_F(\mathbf{r}) - \mathbf{\bar{C}}_F(\mathbf{r})||_2^2\right].
\end{equation}
\newline

\PAR{Airlight Color Supervision}
As discussed in Sec.~\ref{sec:Kohnschmieder}, the model separation allows us to supervise the airlight directly.
We minimize the variance of the predicted $\mathbf{\hat{c}}_p$ with respect to a ground-truth airlight $\mathbf{\bar{c}}_p$ estimated following \cite{tang2014investigating}. The target airlight $\mathbf{\bar{c}}_p$ is computed as follows,
\begin{equation}
\mathbf{\bar{c}}_p(\mathbf{r}) = \frac{z^2(\mathbf{r}) I_F(\mathbf{r}) + \lambda \mathbf{c}_p^0(\mathbf{r})}{z^2(\mathbf{r}) + \lambda}.
\end{equation}
Here $I_F$ is the relative luminance of the hazed image estimated as $I_F(\mathbf{r})= \mathbf{\xi} \cdot \text{lin}(\bar{\mathbf{C}}_F(\mathbf{r}))$, that is a linear combination  of the linearized RGB values \cite{ProposalColorEncoding} obtained by decompanding the color image by applying $\text{lin}(\mathbf{\bar{C}}_F)$ \cite{1999standardRGBColorSpaces}. 
$\mathbf{c}_p^0$ is an initial global constant airlight estimate computed with the dark channel prior following \cite{he2010darkchannelprior}, $z = 1/(l-1)$ and $\lambda$ is a weighting factor. 
The total loss can be written as,
\begin{equation}
L_{A} =\EX_\mathbf{r} \left[ || \mathbf{\hat{\mathbf{c}}}_p(\mathbf{r}) - \mathbf{\bar{\mathbf{c}}}_p(\mathbf{r}) ||_2^2\right].
\end{equation}
\newline

\PAR{Clear Scene Entropy Minimization} 
To regularize $f_c$, we follow \cite{deng2022depth} according to which the rays cast in the scene have peaky unimodal ray termination distributions. Thus, we add a loss to minimize its distribution entropy,
\begin{equation}
\label{eqn:entropyclearloss}
L_{ec} =\EX_{r} \left[ - \sum_i \hat{w}_{c_i}(\mathbf{r}) \cdot \log \left(\hat{w}_{c_i}(\mathbf{r}) \right) \right].
\end{equation}

\PAR{Foggy Scene Entropy Maximization}
Analogously to Eq.~\eqref{eqn:entropyclearloss}, we regularize the density of the scattering media $\hat{\sigma}_p$. Thereby, $\hat{\sigma}_p$ is fitted in semi-supervised fashion during the optimization. This allows us to model fog inhomogeneities, for example close to hot surfaces.
To achieve this goal we apply an entropy-based loss which allows the network $f_p$ to learn a spatially-varying media density. 
Based on the assumption of almost-uniformity for extended fog volume, we enforce that this has to be represented by  the fog volume density $\hat{\sigma}_F$ through maximizing the entropy as follows, 
\begin{equation}
L_{eF} =\EX_r \left[ \sum_i \tilde{\alpha}_{F_i} (\mathbf{r}) \cdot \log( \tilde{\alpha}_{F_i}(\mathbf{r})) \right],
\end{equation}
where $\tilde{\alpha}_{F_i} = \frac{\hat{\alpha}_{F_i}}{ \sum_j \hat{\alpha}_{F_j} } $ and $\hat{\alpha}_{F_i} = 1 - exp((\hat{\sigma}_{p_i} +\hat{\sigma}_{c_i} ) \delta_i )$. \newline
The entropy maximization relies on the scene volume density $\hat{\sigma}_c$ to disentangle both distributions and not only distribute the fog volume density $\hat{\sigma}_F$ throughout the scene. We minimize this loss only for $\sigma_p$, not $\sigma_c$ .
\newline

\PAR{Estimated Depth Supervision}
To supervise the scene representation further we supervise the scene depth $\hat{D}_c$ similar to \cite{roessle2022dense}, through the depth $\bar{D}$ estimated from the stereo sensor setup. Thereby, the depth can be estimated as follows, 
\begin{equation}
    \hat{D}_c(\mathbf{r}) = \sum_{i}^{N} \hat{w}_{c}(\mathbf{r}(t_i)) t_i
\end{equation}
which leads to,
\begin{equation}
\label{eqn:depth_sup}
L_{depth} = \EX_r \left[ || \hat{D}_c(\mathbf{r}) - \bar{D}(\mathbf{r}) ||_2^2 \right]. 
\end{equation}
Stereo depth networks are robust in foggy conditions \cite{depthrobustness} and we find them suitable for supervising the scene NeRF directly. 
\PAR{Total Training Loss}
Combining the five losses, we obtain the following loss formulation, 
\begin{equation}
L_{tot} = \psi_1 L_{rgbF} + \psi_2 L_A + \psi_3 L_{ec} + \psi_4 L_{eF} + \psi_5 L_{depth} .
\end{equation}
Where $\psi_{1,...5}$ are the loss weights, provided in the Supplementary Material.

\begin{figure}[!t]
    \centering
    \includegraphics[width=0.49\textwidth]{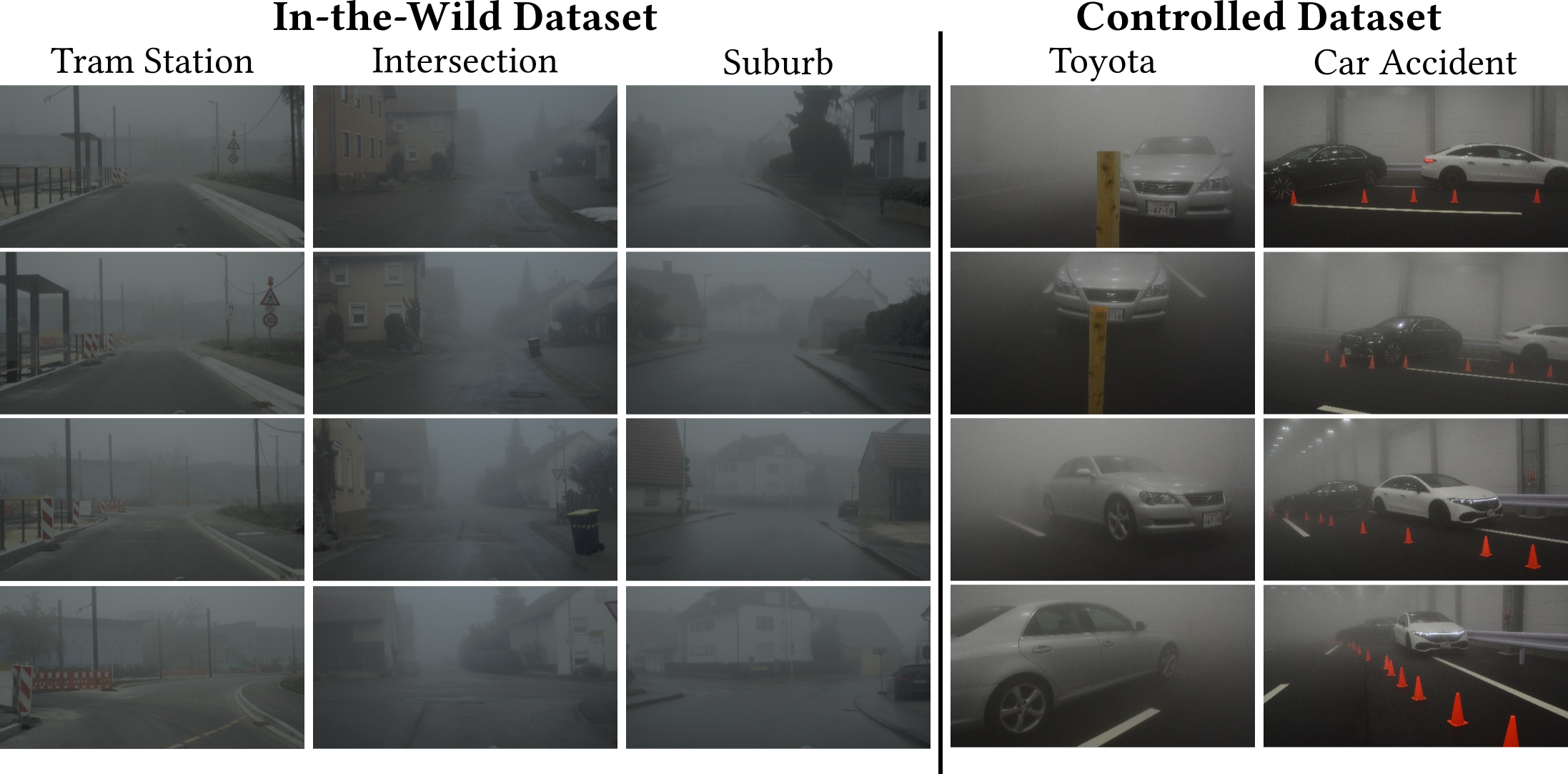}   
    \vspace{-2mm}
    \caption{Examples of our proposed in-the-wild and controlled environment dataset. The figure demonstrates the diversity of the scenes and the fog densities.}
    \label{fig:dataset}
    \vspace{-3mm}
\end{figure}

\subsection{Sampling}\label{sec:Sampling}
We follow the hierarchical volume sampling strategy proposed in \cite{mildenhall2021nerf} to sample the query points for the radiance field networks. However, instead of sampling across the whole volume density we adapt the approach to our decomposed volume densities $\sigma_c$ and $\sigma_F$. As $\sigma_F$ is regularized to be approximately constant across the scene, the re-sampling procedure is not going to be performed using the scene weights $w_F = T_F (1-\exp(\sigma_F \delta))$ but rather using the clear scene $w_c = T_c (1-\exp(\sigma_c \delta) )$. We apply this approach to follow an importance sampling and reconstruct scene objects by sampling close to object boundaries. 

\subsection{Implementation Details}\label{sec:Implementation}
We train for 250'000 steps and a batch size of 4096 rays. As optimizer we use ADAMW \cite{ADAMW} with $\beta_1\,=\,0.9$, $\beta_2\,=\,0.999$, learning rate $5\cdot10^{-4}$ and weight decay factor of $10^{-2}$. Our code implementation is based on Pytorch \cite{pytorch} and we train on four NVIDIA RTX A6000. 
The NeRF MLPs $f_{c_{coarse}}$ and $f_{c_{fine}}$ follow Mildenhall et al.~\cite{mildenhall2021nerf}, while we use fewer hidden layers for $f_p$. The network architecture and other hyper-parameters are listed in the Supplementary <aterial.

\section{Dataset}
\label{sec:dataset}
To evaluate the proposed method, we collect both an automotive in-the-wild foggy dataset and a controlled fog dataset. Example captures illustrating the dataset are shown in Fig.~\ref{fig:dataset}.
In total, we collect 2678 In-the-Wild foggy images throughout nine different scenarios.
The sensor set for the in-the-wild dataset consists of an Aptina-0230 stereo camera pair and a Velodyne HDL64S3D laser scanner. 
Camera poses are estimated with the hierarchical localization approach \cite{sarlin2019coarse, sarlin2020superglue}, a structure from motion pipeline optimized for robustness to changing conditions.
Ground-truth depth data is estimated with the stereo-camera method described in \cite{li2022practical}. The controlled fog dataset is captured in a fog chamber where fog for varying visibilities can be generated.
We capture 903 images in a large-scale fog chamber with clear ground-truth and two different fog densities.
Further ground-truth depths are captured through a Leica ScanStation P30
laser scanner (360°/290° FOV, 1550 nm, with up to 1M points per second, up to 8” angular accuracy, and 1.2mm + 10 parts per million (ppm) range accuracy).
Each point cloud consists of approximately 157M points and we accumulate multiple point clouds from different positions to reduce occlusions and increase resolution.

%% file: sections/results.tex
\input{tables/results_table}
\input{tables/results_table_controlled}
\section{Assessment}
Next, we validate the proposed method by ablating the different components, confirming their effectiveness, assessing the quality of scene reconstructions in foggy scenes, and the decomposition into dehazed scene content.
\subsection{Ablation}
\input{tables/ablation_table}
\begin{figure}[!t]
    \centering
    \includegraphics[width=0.49\textwidth]{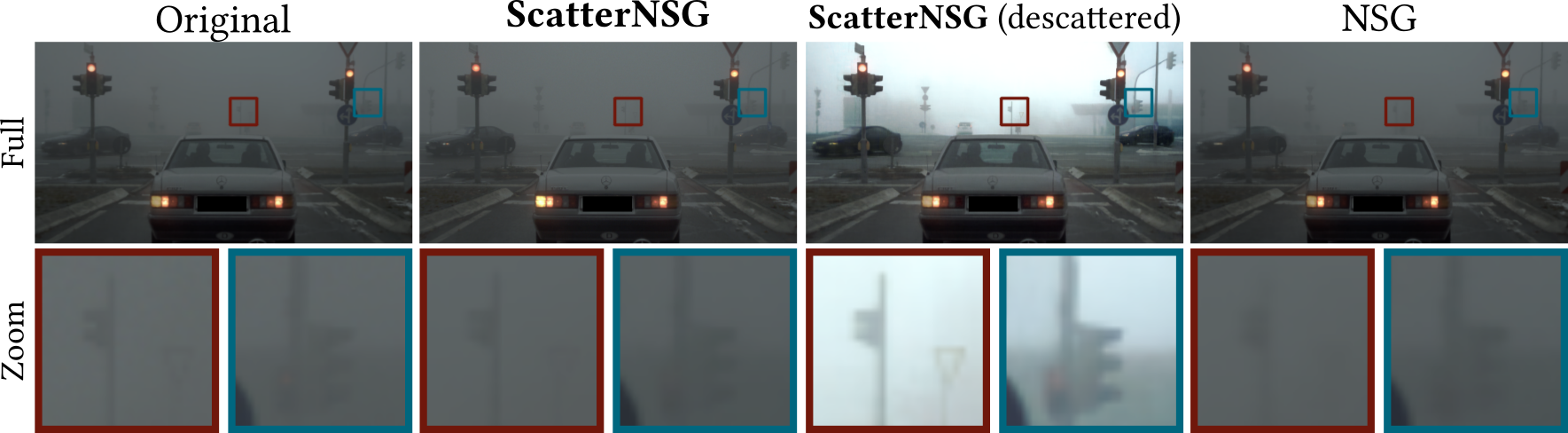}   
    \vspace{-2mm}
    \caption{Qualitative comparison between Neural Scene Graphs (NSG) \cite{ost2021NSG} and its combination with our proposed network (ScatterNSG).}
    \label{fig:figure_nsg}
    \vspace{-1mm}
\end{figure}

In order to assess the role and contribution of the different components of our framework, we conduct an ablation study whose results are presented in Tab.~\ref{tab:ablation}. We consider as starting point the \textit{"NeRF"} \cite{mildenhall2021nerf}.  Its PSNR on the In-the-Wild subset dataset is 39.06 dB. By adding the depth supervision Eq.~\eqref{eqn:depth_sup} and Eq.~\eqref{eqn:entropyclearloss} the model's PSNR improves by 0.97 dB.
However, as shown in Fig. \ref{fig:arch}, a NeRF trained in such a way does not produce a physically accurate representation of the scene, as it does not model the participating media present in the scene, but rather represents it as a clear scene.
Adding the scattering media $f_p$ in \textit{"ScatterNerf w/o cleared sampling"}, together with its two losses $L_A$ and $L_{eF}$ leads to a PSNR of 40.7 dB. Finally, adding the sampling strategy in the full \textit{"ScatterNerf"} helps the model achieve better results, with a PSNR of 41.45 dB, summing up to a 6.12\% PSNR increase over the baseline. \newline
Here, we also analyze an additional model, in which a Koschmieder model is added to the NeRF output in \textit{"Nerf+Koschmieder"} of Tab.~\ref{tab:ablation}. Qualitative results of the models used in the ablation are shown in the supplementary material.

\begin{table}[!t]
    \centering
    \resizebox{.49\textwidth}{!}{
    \input{tables/table_fid.tex}
    }
    \vspace{-2mm}
    \caption{\small Quantitative dehazing comparison on In-the-Wild dataset with FiD score and ond controlled dataset with PSNR. The best results in each category are in \textbf{bold} and the second best are
    \underline{underlined}.}
    \label{tab:fid_score}
    \vspace{-1mm}
\end{table}

\begin{figure}[!t]
    \centering
    \includegraphics[width=0.49\textwidth]{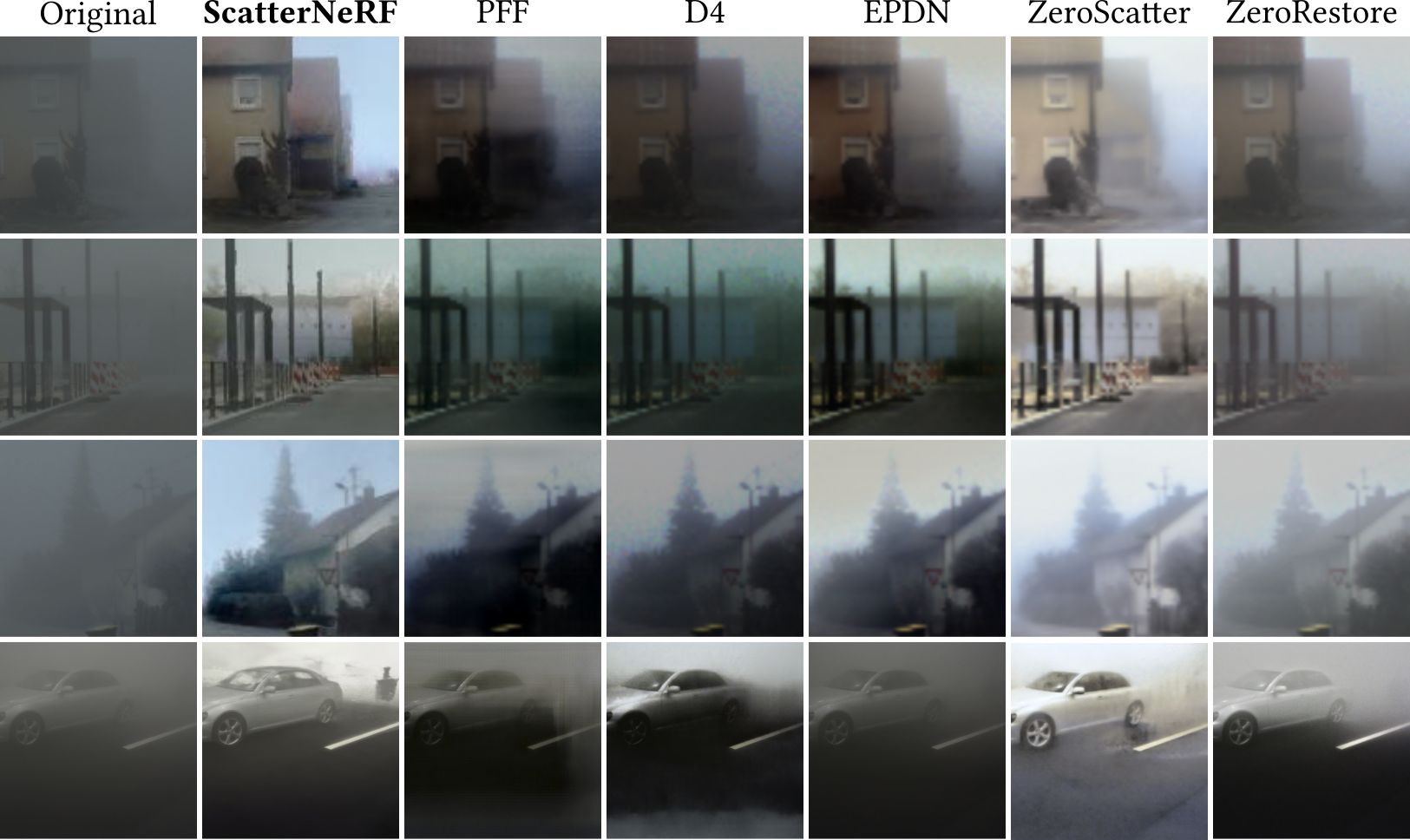}   
    \vspace{-5mm}
    \caption{\small Qualitative comparison of dehazed images on real-world automotive measurmeents. The proposed ScatterNeRF enables enhanced contrast and visibility compared to state-of-the-art descattering methods.}
    \label{fig:qual_defogged_automotive_results}
    \vspace{-3mm}
\end{figure}

\begin{figure*}[!t]
    \centering
    \includegraphics[width=0.99\textwidth]{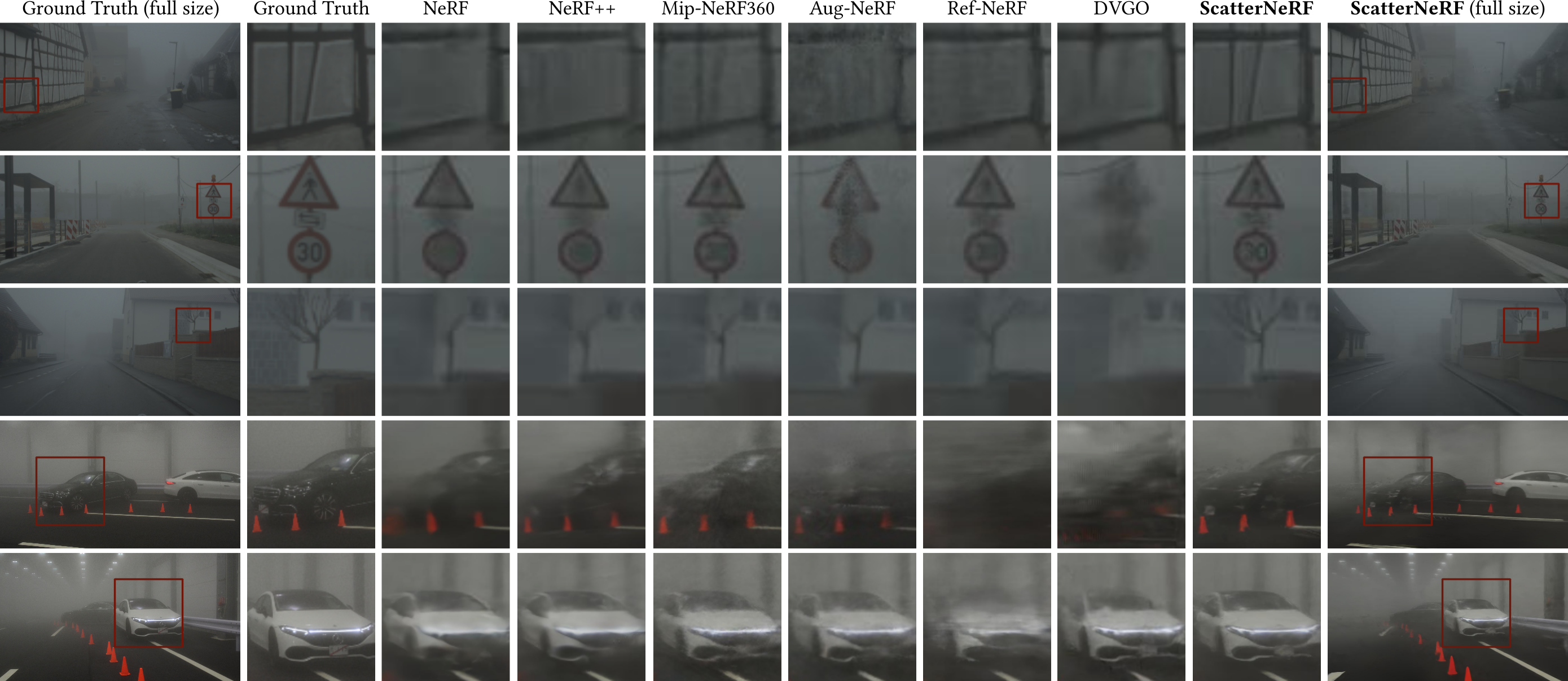}   
    \vspace{-2mm}
    \caption{\small Qualitative comparisons of the reconstruction of foggy scenes with ScatterNeRF and state-of-the-art neural rendering mehtods. ScatterNeRF is able to represent the participating media much better than existing rendering methods.}
    \label{fig:qual_results_foggy_reconstruction}
    \vspace{-1mm}
\end{figure*}

\subsection{Foggy Scene Reconstruction}
We compare our proposed method with NeRF \cite{mildenhall2021nerf}, Mip-NeRF \cite{barron2021mip}, DVGO \cite{SunSC22}, Plenoxel \cite{fridovich2022plenoxels}, Ref-NeRF \cite{verbin2022refNeRF}, two methods for unbounded scenes \cite{zhang2020nerf++, barron2022mip} and two NeRFs with auxiliary regularization \cite{roessle2022dense, chen2022augnerf}. 
Qualitative results are presented in Fig.~\ref{fig:qual_results_foggy_reconstruction}. The baseline methods struggle with object edges and fine structures, which our method is able to reconstruct for novel views. Our method is the only approach able to reconstruct the edges cleanly achieving to the highest PSNR scores for in-the-wild captures and fog chamber captures. On average, the proposed method improves on average by 2.13~dB the PSNR of NeRF and by 1.64~dB to the next best model. For single sequences with high fog densities, improvements in the reconstruction of up to 2.39~db are measurable. For the SSIM metric it outperforms all other approaches except on the light foggy scene dubbed "car accident" where it seconds \cite{fridovich2022plenoxels} and matches \cite{barron2021mip}. \newline

\subsection{Generalizability}
As our proposed method does not require major changes to the sampling procedure, rendering or scene representation architecture, it can be easily integrated with existing neural rendering methods.
To demonstrate this, we extend NSG \cite{ost2021NSG} to foggy scenes and integrate the decoupling of scene and scattering media. More details are given in the supplemental material. A qualitative comparison between the base model and the one enhanced with our framework is shown in Fig.~\ref{fig:figure_nsg}, where it is possible to observe both the finer details at long distances in addition to the possibility to remove the scattering media and better reconstruct the vehicle leaving the scene. Quantitatively this improvement is reflected in an improvement of scene reconstruction PSNR, increasing from 29.78 dB to 30.67 dB. 
This also validates that it is possible to employ our framework \emph{also in the presence of moving objects}, which is a common occurrence in automotive scenes.
We discuss in the supplementary material other examples of the integration of ScatterNeRF in state-of-the-art neural radiance fields to augment their scope.

\subsection{Dehazing}
We quantify the ability of ScatterNeRF to learn a disentangle representation between objects and scattering media, and hence to render the corresponding clear scene which is effectively dehazing the foggy image.
To this end, we compare against the state-of-the art dehazing methods PFF \cite{PFF}, D4 \cite{d4Unpaired}, EPND \cite{EPDN}, ZeroScatter \cite{shi2021zeroscatter} and ZeroRestore \cite{zerorestore}. 
For the in-the-wild scenes, we rely on the FiD score \cite{heuselFID2017} to evaluate the quality of the dehazing.
As ground-truth data is unavailable, this score allows us to compare the dehazed sequences with a similar clear-weather scene collected with the automotive setup described in Sec.~\ref{sec:dataset}.
For the controlled environment dataset the PSNR can directly be evaluated by warping a nearby clear-weather image to the respective foggy image.
In contrast to other dehazing methods operating on single frames, ScatterNeRF learns a consistent representation of the entire sequence.
As a result, the dehazing is consistent across consecutive frames, whereas the baseline methods often are affected by flickering effects.
This is also reflected in the quantitative results in Tab.~\ref{tab:fid_score}.
Here, ScatterNeRF outperforms the baseline methods on almost all sequences indicating a consistent dehazing across the entire sequence.
Furthermore, the qualitative results in Fig.~\ref{fig:qual_defogged_automotive_results} from the controlled fog chamber dataset reveal another strength of ScatterNeRF:
ScatterNeRF is the only method able to reconstruct the cart behind the vehicle as it can leverage information from the learned representation of the entire sequence. This results in a higher PSNR indicating an improved dehazing on the respective sequence.
Additionally, as evident in Fig.~\ref{fig:qual_defogged_automotive_results}, ScatterNeRF achieves visually enhanced contrast compared to existing dehazing methods.

%% file: tables/results_table.tex
\begin{table*}[!tp]
    \footnotesize
    \setlength{\tabcolsep}{4pt} % general space between cols (6pt standard)
    \setlength\extrarowheight{2pt}
    \centering
    \resizebox{.99\textwidth}{!}{
    \begin{tabular}{@{}lccccccccccccccc@{}}
            \toprule
            & \multicolumn{3}{|c|}{\textbf{Tram Station}} & \multicolumn{3}{|c|}{\textbf{Farm}} & \multicolumn{3}{|c|}{\textbf{Intersection}} & \multicolumn{3}{|c|}{\textbf{Suburb}} & \multicolumn{3}{|c|}{\textbf{Speed Control}}\\
            \textbf{\textsc{Method}} & LPIPS $\downarrow$ & PSNR $\uparrow$ & SSIM $\uparrow$ & LPIPS $\downarrow$ & PSNR $\uparrow$ & SSIM $\uparrow$ & LPIPS $\downarrow$ & PSNR $\uparrow$ & SSIM $\uparrow$ & LPIPS $\downarrow$ & PSNR $\uparrow$ & SSIM $\uparrow$ & LPIPS $\downarrow$ & PSNR $\uparrow$ & SSIM $\uparrow$ \\ 
            \midrule
            \input{tables/automotive_results_row1} \\
            \midrule
            & \multicolumn{3}{|c|}{\textbf{Road Fork}} & \multicolumn{3}{|c|}{\textbf{Adds}} & \multicolumn{3}{|c|}{\textbf{Construction}} & \multicolumn{3}{|c|}{\textbf{Countryside}} & \multicolumn{3}{|c|}{\textbf{Average In-the-Wild}}\\
            \textbf{\textsc{Method}} & LPIPS $\downarrow$ & PSNR $\uparrow$ & SSIM $\uparrow$ & LPIPS $\downarrow$ & PSNR $\uparrow$ & SSIM $\uparrow$ & LPIPS $\downarrow$ & PSNR $\uparrow$ & SSIM $\uparrow$ & LPIPS $\downarrow$ & PSNR $\uparrow$ & SSIM $\uparrow$ &  LPIPS $\downarrow$ & PSNR $\uparrow$ & SSIM $\uparrow$ \\ 
            \midrule
            \input{tables/automotive_results_row2} \\
            \bottomrule
    \end{tabular}
    }
    \vspace*{-3pt}
    \caption{\label{tab:results_foggy_reconstruction}\small Quantitative comparison of the proposed ScatterNeRF and state-of-the-art methods on In-the-Wild sequences. Best results in each category are in \textbf{bold} and second best are
    \underline{underlined}. Last column in the second row presents the average over all sequences.
	\vspace*{-1mm}
}
\end{table*}

%% file: tables/automotive_results_row1.tex
Aug-NeRF \cite{chen2022augnerf} & 0.26 & 39.18 & 0.964 & 0.316 & 39.83 & 0.968 & 0.275 & 40.38 & 0.971 & 0.259 & 40.18 & 0.977 & 0.277 & 40.54 & 0.974 \\ 
DS-NeRF \cite{deng2022depth} & 0.265 & 39.99 & 0.967 & 0.332 & \underline{41.27} & 0.97 & 0.273 & \underline{42.19} & \underline{0.977} & 0.271 & \underline{42.89} & \underline{0.978} & 0.281 & \underline{42.8} & 0.976 \\ 
DVGO \cite{SunSC22} & 0.404 & 35.46 & 0.923 & 0.441 & 37.49 & 0.93 & 0.4 & 39.35 & 0.955 & 0.36 & 39.68 & 0.963 & 0.382 & 39.16 & 0.958 \\ 
Mip-NeRF \cite{barron2021mip} & 0.333 & 40.29 & 0.964 & 0.337 & 40.52 & 0.968 & 0.271 & 41.72 & 0.975 & 0.27 & 41.23 & 0.977 & 0.28 & 41.72 & 0.975 \\ 
Mip-NeRF360 \cite{barron2022mip} & \underline{0.222} & \underline{40.92} & \underline{0.971} & \textbf{0.28} & 41.25 & \textbf{0.973} & \textbf{0.228} & 42.17 & 0.976 & \underline{0.244} & 41.91 & 0.977 & \textbf{0.247} & 42.2 & \underline{0.977} \\ 
NeRF \cite{mildenhall2021nerf} & 0.278 & 39.06 & 0.964 & 0.354 & 40.86 & 0.968 & 0.293 & 41.84 & 0.974 & 0.271 & 41.6 & \underline{0.978} & 0.293 & 42.39 & 0.976 \\ 
NeRF++ \cite{zhang2020nerf++} & 0.288 & 39.54 & 0.962 & 0.36 & 41.07 & 0.966 & 0.306 & 42.11 & 0.972 & 0.293 & 42.07 & 0.976 & 0.304 & 42.41 & 0.974 \\ 
Plenoxels \cite{fridovich2022plenoxels} & 0.476 & 30.9 & 0.91 & 0.498 & 32.28 & 0.939 & 0.478 & 29.93 & 0.934 & 0.486 & 28.29 & 0.928 & 0.489 & 29.34 & 0.934 \\ 
Ref-NeRF \cite{verbin2022refNeRF} & 0.4 & 36.64 & 0.94 & 0.377 & 39.85 & 0.966 & 0.377 & 40.53 & 0.964 & 0.343 & 40.47 & 0.972 & 0.366 & 41.13 & 0.967 \\ 
\textbf{ScatterNeRF} & \textbf{0.22} & \textbf{41.45} & \textbf{0.975} & \underline{0.299} & \textbf{42.57} & \underline{0.972} & \underline{0.235} & \textbf{44.58} & \textbf{0.98} & \textbf{0.234} & \textbf{44.44} & \textbf{0.981} & \underline{0.253} & \textbf{43.88} & \textbf{0.978} 

%% file: tables/automotive_results_row2.tex
Aug-NeRF \cite{chen2022augnerf} & 0.248 & 41.29 & \underline{0.979} & 0.242 & 41.75 & 0.981 & 0.302 & 38.99 & 0.964 & 0.286 & 40.74 & 0.978 & 0.274 & 40.32 & 0.973 \\ 
DS-NeRF \cite{deng2022depth} & 0.273 & \underline{42.59} & 0.977 & 0.252 & 43.22 & \underline{0.982} & 0.318 & 40.94 & 0.964 & 0.29 & \underline{43.6} & 0.98 & 0.284 & \underline{42.16} & 0.975 \\ 
DVGO \cite{SunSC22} & 0.376 & 38.51 & 0.947 & 0.349 & 40.14 & 0.966 & 0.376 & 38.92 & 0.95 & 0.368 & 40.55 & 0.968 & 0.384 & 38.81 & 0.951 \\ 
Mip-NeRF \cite{barron2021mip} & 0.272 & 41.88 & 0.976 & 0.25 & 42.83 & 0.981 & 0.321 & 40.2 & 0.963 & 0.291 & 43.08 & 0.98 & 0.292 & 41.5 & 0.973 \\ 
Mip-NeRF360 \cite{barron2022mip} & \underline{0.235} & 42.37 & 0.978 & \textbf{0.225} & 42.55 & 0.981 & \underline{0.268} & \underline{41.53} & \underline{0.969} & \textbf{0.241} & 42.97 & \underline{0.981} & \textbf{0.243} & 41.99 & \underline{0.976} \\ 
NeRF \cite{mildenhall2021nerf} & 0.287 & 41.84 & 0.975 & 0.256 & 43.67 & \underline{0.982} & 0.332 & 40.36 & 0.962 & 0.305 & 43.39 & 0.979 & 0.297 & 41.67 & 0.973 \\ 
NeRF++ \cite{zhang2020nerf++} & 0.292 & 42.38 & 0.975 & 0.267 & \underline{43.82} & 0.98 & 0.336 & 40.69 & 0.962 & 0.312 & 43.56 & 0.978 & 0.306 & 41.96 & 0.972 \\ 
Plenoxels \cite{fridovich2022plenoxels} & 0.481 & 29.85 & 0.933 & 0.456 & 29.72 & 0.943 & 0.497 & 29.49 & 0.922 & 0.473 & 29.47 & 0.938 & 0.482 & 29.92 & 0.931 \\ 
Ref-NeRF \cite{verbin2022refNeRF} & 0.382 & 39.61 & 0.962 & 0.376 & 40.81 & 0.967 & 0.365 & 40.04 & 0.96 & 0.339 & 42.81 & 0.976 & 0.369 & 40.21 & 0.964 \\ 
\textbf{ScatterNeRF} & \textbf{0.231} & \textbf{44.45} & \textbf{0.981} & \underline{0.228} & \textbf{45.27} & \textbf{0.983} & \textbf{0.255} & \textbf{42.96} & \textbf{0.973} & \underline{0.265} & \textbf{44.64} & \textbf{0.982} & \underline{0.247} & \textbf{43.8} & \textbf{0.978}

%% file: tables/results_table_controlled.tex
\begin{table}[!tp]
	\footnotesize
	\setlength{\tabcolsep}{6pt} % general space between cols (6pt standard)
	\setlength\extrarowheight{2pt}
	\centering
	\resizebox{.49\textwidth}{!}{
		\begin{tabular}{@{}lcccccc@{}}
			\toprule
			& \multicolumn{3}{|c|}{\textbf{Toyota}} & \multicolumn{3}{|c|}{\textbf{Car Accident}} \\
			\textbf{\textsc{Method}} & LPIPS $\downarrow$ & PSNR $\uparrow$ & SSIM $\uparrow$ & LPIPS $\downarrow$ & PSNR $\uparrow$ & SSIM $\uparrow$  \\  
			\midrule
			\input{tables/controlled_results} \\
			\bottomrule
		\end{tabular}
	}
	\caption{\label{tab:results_controlled}\small Quantitative comparison of the proposed ScatterNeRF and state-of-the-art methods on controlled scenes. Best results in each category are in \textbf{bold} and second best are \underline{underlined}.}
\end{table}

%% file: tables/controlled_results.tex
Aug-NeRF \cite{chen2022augnerf} & 0.521 & 29.81 & 0.934 & \textbf{0.48} & 25.08 & 0.822 \\ 
DS-NeRF \cite{deng2022depth} & 0.535 & 29.52 & 0.926 & \underline{0.502} & \underline{26.71} & 0.853 \\ 
DVGO \cite{SunSC22} & 0.613 & 22.65 & 0.801 & 0.607 & 22.22 & 0.847 \\ 
Mip-NeRF \cite{barron2021mip} & 0.513 & 30.42 & 0.937 & 0.611 & 24.88 & 0.87 \\ 
Mip-NeRF360 \cite{barron2022mip} & 0.53 & \underline{30.59} & \textbf{0.938} & 0.623 & 26.59 & \underline{0.878} \\ 
NeRF \cite{mildenhall2021nerf} & \underline{0.505} & 29.81 & 0.937 & 0.614 & 24.97 & 0.877 \\ 
NeRF++ \cite{zhang2020nerf++} & 0.507 & 30.0 & 0.937 & 0.611 & 25.47 & \textbf{0.879} \\ 
Plenoxels \cite{fridovich2022plenoxels} & 0.582 & 20.48 & 0.872 & 0.607 & 22.07 & 0.852 \\ 
Ref-NeRF \cite{verbin2022refNeRF} & \textbf{0.498} & 29.11 & 0.933 & 0.616 & 24.7 & 0.864 \\ 
\textbf{ScatterNeRF} & \underline{0.505} & \textbf{30.82} & \textbf{0.938} & 0.509 & \textbf{27.42} & \underline{0.878} 
%ScatterNeRF & 0.505 & 30.82 & 0.937 & 0.509 & 27.42 & 0.878
%\textbf{ScatterNeRF} & 0.438 & 27.95 & 0.8905 & -1.0 & -1.0 & -1.0 & 0.495 & 23.11 & 0.784 & -1.0 & -1.0 & -1.0 

% \textbf{ScatterNeRF} & 0.438 & 27.95 & 0.8905 & -1.0 & -1.0 & -1.0 & 0.495 & 23.11 & 0.784 & -1.0 & -1.0 & -1.0 

%% file: tables/ablation_table.tex
\begin{table}[!tp]
    \footnotesize
    \setlength{\tabcolsep}{2pt} % general space between cols (6pt standard)
    \setlength\extrarowheight{2pt}
    \centering
    % \resizebox{.24\textwidth}{!}{
    \begin{tabular}{@{}lccccccccc@{}}
            \toprule
             & LPIPS & PSNR & SSIM \\ 
            \midrule
            NeRF & 0.278 & 39.06 & 0.964 \\
            NeRF + Koschmieder & 0.281 & 39.54 & 0.960\\                        
            NeRF + Depth \cite{roessle2022dense} & 0.290 & 40.03 & 0.972\\
            ScatterNeRF w/o cleared sampling & 0.219 & 40.70 & 0.975 \\
            ScatterNeRF & 0.22 & 41.45 & 0.975\\
			% \multirow{6}{*}{\rotatebox[origin=l]{90}{\parbox[c]{1.6cm}{\centering \textbf{\small{last}}}}}
			% \input{ScatterNerf/tables/results_last}\\
 		% 	\midrule
 		% 	\multirow{6}{*}{\rotatebox[origin=l]{90}{\parbox[c]{1.6cm}{\centering \textbf{\small{random}}}}}
			% \input{ScatterNerf/tables/results_random}\\
			% \midrule
   %          \multirow{6}{*}{\rotatebox[origin=l]{90}{\parbox[c]{1.6cm}{\centering \textbf{\small{left/right}}}}}
			% \input{ScatterNerf/tables/results_left_right}\\
			% \midrule
            \bottomrule
    \end{tabular}
    % }
    \vspace*{0pt}
    \caption{\small Ablation study of the ScatterNerf contribution for a subset of the In-the-Wild dataset. 
	\vspace*{0mm}
}
\label{tab:ablation}
\end{table}

%% file: tables/table_fid.tex
\begin{tabular}{l|l|cccccc}
\toprule
\textsc{\textbf{Sequence}} && \textbf{ScatterNerf} & PFF & D4 & EPDN & ZeroScatter & ZeroRestore \\
\midrule
Tram Station & \multirow{9}{*}{\rotatebox[origin=l]{90}{\parbox[c]{0.8cm}{\centering \footnotesize{FiD $\downarrow$}}}} & 348.72 & 376.39 & \underline{342.10} & 345.66 & 355.27 & \textbf{340.59} \\
Farm && \textbf{319.95} & 404.89 & 360.53 & 347.85 & 413.08 & \underline{359.53} \\
Intersection && \textbf{349.57} & 387.09 & \underline{358.63} & 353.27 & 397.71 & 366.82 \\
Suburb && \textbf{283.43} & 405.37 & \underline{299.57} & 301.63 & 347.38 & 350.11 \\
Speed Control && \textbf{270.86} & 409.94 & 332.78 & \underline{328.08} & 358.97 & 339.19 \\
Road Fork && \textbf{284.69} & 419.08 & 318.84 & \underline{314.66} & 355.30 & 325.45 \\
Adds && \textbf{322.19} & 441.33 & \underline{328.59} & 330.71 & 380.70 & 342.91 \\
Construction && 329.22 & 381.51 & \underline{318.71} & \textbf{315.16} & 339.66 & \textbf{310.11} \\
Countryside && \textbf{289.56} & 435.71 & 326.05 & 321.35 & 337.88 & \underline{321.29} \\
\midrule
\noalign{\vskip 0.1cm}
Toyota & \multirow{2}{*}{\rotatebox[origin=l]{90}{\parbox[c]{0.8cm}{\centering \footnotesize{PSNR~$\uparrow$}}}} & \textbf{13.47} & 12.13 & 11.74 & 11.62 & 13.39 & \underline{13.41} \\
Car Accident && \underline{11.66} & 10.76 & 10.32 & 10.04 & 9.28 & \textbf{12.02} \\
 \noalign{\vskip 0.1cm}    

%&& &&&&&& \\
\bottomrule
\end{tabular}

%% file: sections/conclusion.tex
\section{Conclusion}
We introduce ScatterNeRF, a neural rendering method that represents foggy scenes with disentangled representations of the participating media and the scene.
We model light transport in the presence of scattering with disentangled volume rendering, separately modeling the clear scene and fog, and introduce a set of physics-based losses designed to enforce the division between media and scene. 
Extensive experiments with both in-the-wild and controlled scenario measurements validate the proposed approach. 
We demonstrate that ScatterNeRF is capable of rendering the learned scene without scattering media and can be hence used to alter or remove the haze from a sequence video, reaching quality comparable to state-of-the-art image dehazing algorithms -- solely by fitting image observations of the scene, without any forward neural network for dehazing or denoising.